\documentclass[10pt,twocolumn,letterpaper]{article}

\usepackage{iccv}              

\usepackage{amssymb}
\usepackage{amsmath}
\usepackage{multirow}
\usepackage{booktabs}
\usepackage{makecell}
\usepackage{colortbl}
\usepackage{booktabs}
\usepackage{pifont}
\usepackage{stfloats}

\definecolor{iccvblue}{rgb}{0.21,0.49,0.74}
\usepackage[pagebackref,breaklinks,colorlinks,allcolors=iccvblue]{hyperref}

\def\paperID{11926} 
\def\confName{ICCV}
\def\confYear{2025}

\title{ForgeLens: Data-Efficient Forgery Focus for Generalizable \\ Forgery Image Detection}

\author{ 
\textbf{Yingjian Chen},
\textbf{Lei Zhang},
\textbf{Yakun Niu\thanks{Corresponding author}},
\\
Henan Key Laboratory of Big Data Analysis and Processing, Henan University
\\
{\tt\small yingjianchen@henu.edu.cn, zhanglei@henu.edu.cn, ykniu@henu.edu.cn}
}

\begin{document}
\maketitle
\begin{abstract}
 The rise of generative models has raised concerns about image authenticity online, highlighting the urgent need for a detector that is (1) highly generalizable, capable of handling unseen forgery techniques, and (2) data-efficient, achieving optimal performance with minimal training data, enabling it to counter newly emerging forgery techniques effectively. To achieve this, we propose \textbf{\textit{ForgeLens}}, a data-efficient, feature-guided framework that incorporates two lightweight designs to enable a frozen network to focus on forgery-specific features. First, we introduce the Weight-Shared Guidance Module (WSGM), which guides the extraction of forgery-specific features during training. Second, a forgery-aware feature integrator, FAFormer, is used to effectively integrate forgery information across multi-stage features. ForgeLens addresses a key limitation of previous frozen network-based methods, where general-purpose features extracted from large datasets often contain excessive forgery-irrelevant information. As a result, it achieves strong generalization and reaches optimal performance with minimal training data. Experimental results on 19 generative models, including both GANs and diffusion models, demonstrate improvements of 13.61\% in Avg.Acc and 8.69\% in Avg.AP over the base model. Notably, ForgeLens outperforms existing forgery detection methods, achieving state-of-the-art performance with just 1\% of the training data. Our code is available at \url{https://github.com/Yingjian-Chen/ForgeLens}.
\end{abstract}

\section{Introduction}
\label{sec:intro}
Recently, the rapid rise of AI-generated content (AIGC) has attracted significant public interest. Especially in the field of image synthesis, society is facing unprecedented challenges due to the rapid development of technology. The widespread adoption of advanced techniques like Generative Adversarial Networks (GANs)~\cite{goodfellow2014generative} and diffusion models~\cite{rombach2022high} has made it effortless to produce highly realistic fake images. This technological progress has blurred the line between real and synthetic visuals, marking the end of the era where "seeing is believing." These convincingly realistic images are becoming increasingly difficult to distinguish with human vision alone, posing significant risks to societal security. Consequently, effectively identifying such synthetic images has become an urgent concern.

\begin{figure}[t]
    \centering
    \includegraphics[width=1.0\linewidth]{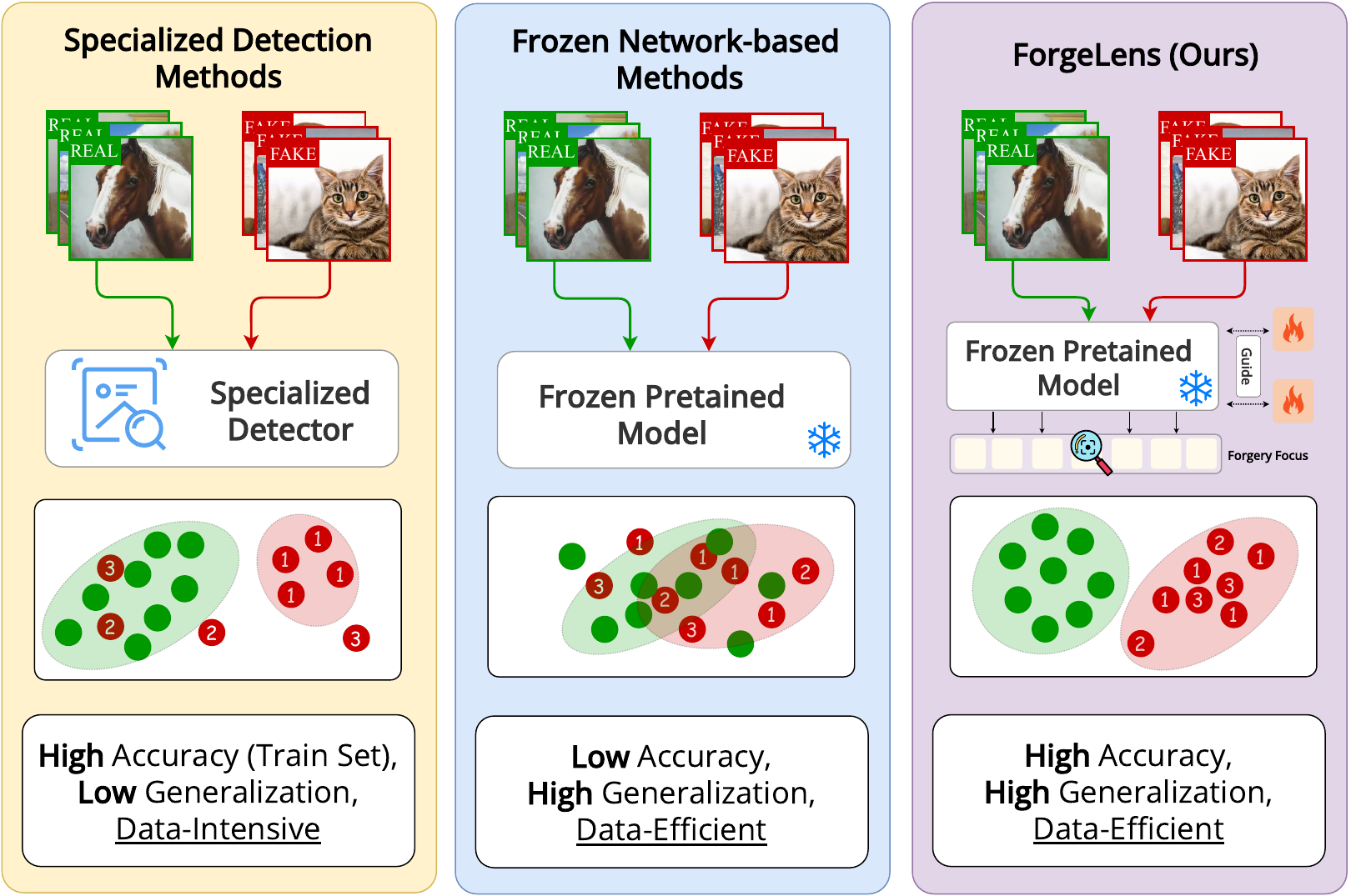}
    \caption{Comparison of forgery image detection methods. Existing \textbf{Specialized Detection Methods} achieve high accuracy on seen datasets but demonstrate limited effectiveness on unseen data. \textbf{Frozen Network-based Methods} exhibit high generalization but lack detection accuracy. In contrast, our \textbf{ForgeLens} achieves high generalization while maintaining superior detection accuracy. The \textcolor[HTML]{009900}{green dots} represent real images, and the \textcolor[HTML]{CC0000}{red dots} represent fake images, with 1 (training set), 2, and 3 showing fake images generated by different models.}
\label{fig: overview}
\end{figure}

\begin{figure*}[t]
    \centering
    \includegraphics[width=1.0\linewidth]{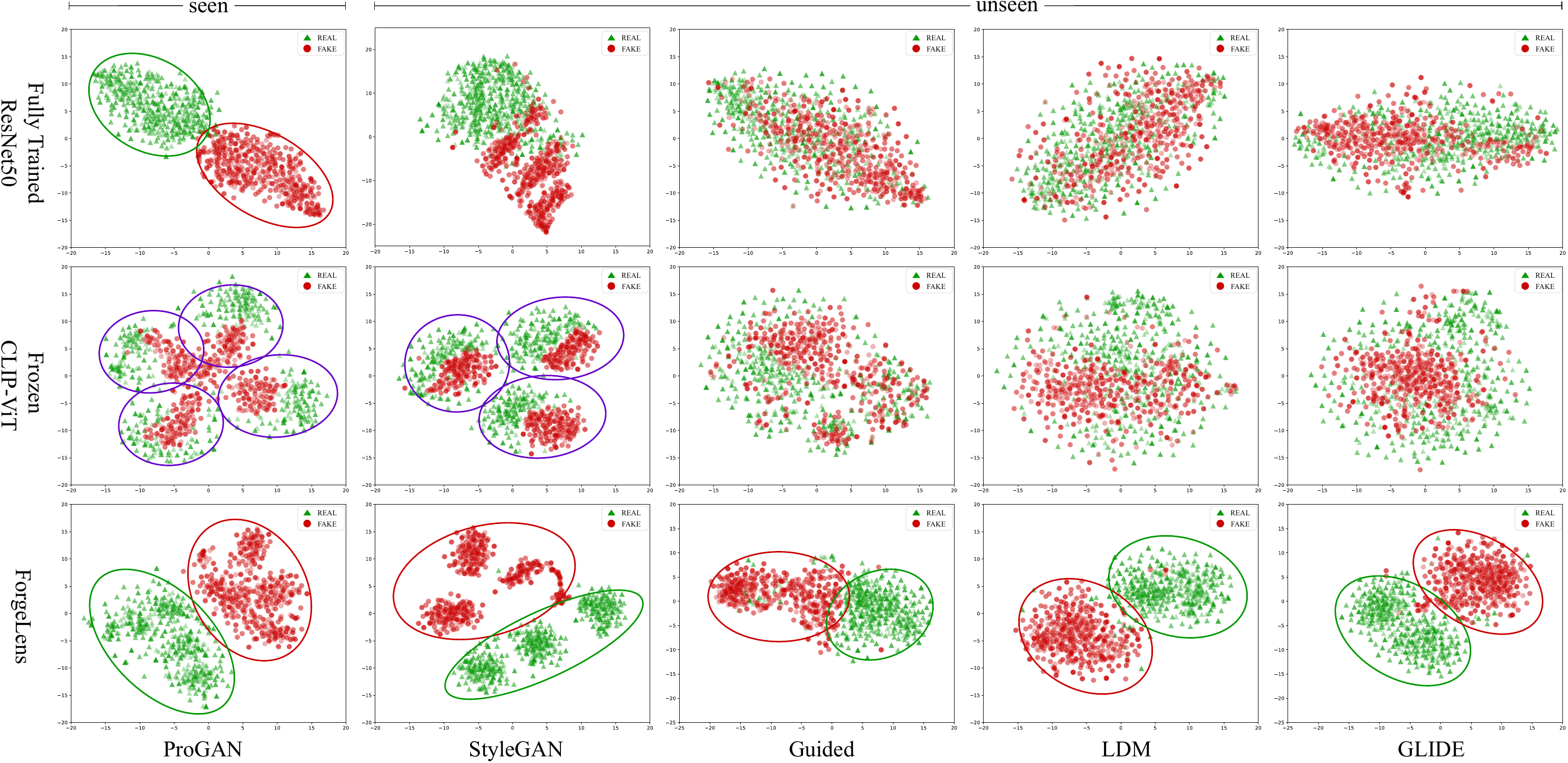}
    \caption{Feature space visualization for GANs (ProGAN~\cite{karras2017progressive}, StyleGAN~\cite{karras2019style}) and diffusion models (Guided~\cite{dhariwal2021diffusion}, LDM~\cite{rombach2022high}, GLIDE~\cite{nichol2021glide}) using t-SNE~\cite{van2008visualizing}. We compare a classic ResNet50 model~\cite{he2016deep} trained from scratch, the base model CLIP-ViT~\cite{radford2021learning}, and our method ForgeLens, all trained on 4-class ProGAN data. Each set contains a random sample of 1k real and 1k fake images. Purple circles represent clusters corresponding to image categories, such as the four categories in ProGAN (car, cat, chair, horse).}
    \label{fig: tsne}
\end{figure*}

In contrast to traditional methods~\cite{farid2016photo, mccloskey2018detecting, mccloskey2019detecting} that rely on handcrafted feature extraction and algorithms, which often suffer from limited detection performance, current approaches have shifted toward deep learning-based techniques, achieving remarkable success in detecting synthetic images. Existing specialized detection methods~\cite{wang2020cnn, deng2023new, ju2023glff, tan2024frequency} design deep learning networks to detect artifacts in the spatial or frequency domains of synthetic images (Figure \ref{fig: overview}, left). These methods achieve high detection accuracy when detecting images generated by seen generative models. However, since the images generated by different generative models exhibit unique artifacts~\cite{corvi2023intriguing, ojha2023towards}, these specialized detection models, which are fully trained from scratch, tend to overfit the training set. As a result, their performance drops significantly when dealing with unseen synthetic images, indicating their limited generalization. In addition, these methods typically require a large amount of training data to achieve optimal performance.

To address this, frozen network-based methods~\cite{ojha2023towards, koutlis2024leveraging} use pre-trained networks for extracting general-purpose image features, which are suitable for various downstream tasks. During training, the network remains frozen, and only a linear classifier is trained (Figure~\ref{fig: overview}, middle). This prevents overfitting and ensures high generalization. However, general-purpose image features often contain excessive forgery-irrelevant information, making it difficult for the classifier to distinguish fake from real images, resulting in limited detection accuracy.

To verify our earlier statements, we visualize the image features extracted by the classic ResNet50 model (used as a representative of fully trained specialized methods) and frozen CLIP-ViT, as shown in Figure \ref{fig: tsne}. The features extracted by ResNet50 (first row) exhibit clear clustering for real and fake images on seen ProGAN data. However, for unseen data, the extracted features become less discriminative. For frozen CLIP-ViT, as mentioned earlier, the extracted features (second row) are general-purpose and not optimized for forgery detection. This is evident in the clustering of ProGAN and StyleGAN data, where features are grouped by image category rather than by real or fake images. This demonstrates CLIP-ViT’s limited ability to distinguish between real and forged images.

Based on the above analysis, we aim to design a forgery image detector for real-world scenarios. It should have: (1) \textbf{strong generalization} to detect images generated by various models, including those generated by previously unseen models, while maintaining high detection accuracy; (2) \textbf{minimal training data requirements}, allowing quick adaptation to new forgery techniques. Therefore, our motivation is: (1) guiding the frozen pre-trained network to focus on forgery-specific features, ensuring high generalization while improving detection accuracy, and (2) introducing as few trainable parameters as possible, enabling the detector to achieve optimal performance with minimal training data. 

To achieve this, we propose \textbf{\textit{ForgeLens}} (Figure \ref{fig: overview}, right), a feature-guided framework that uses a lightweight, trainable Weight-Shared Guidance Module (WSGM) to guide the frozen, pre-trained CLIP-ViT in focusing forgery-specific features. Additionally, a forgery-aware feature integrator, FAFormer, is designed to refine forgery information across multi-stage features, enabling the model to preserve the high generalization of large-scale pre-trained weights while effectively focusing on forgery detection. Through this approach, our method suppresses forgery-irrelevant information in the general-purpose image features extracted by the original CLIP-ViT, thereby obtaining highly discriminative feature representations (Figure \ref{fig: tsne}, third row). Empirically, we find that our method, trained solely on ProGAN-generated images, exhibits strong generalization on the UniversalFakeDetect dataset~\cite{ojha2023towards}, which contains images from various generative models, including both GANs and diffusion models. It achieves a high average detection accuracy of 94.99\%, demonstrating its effectiveness in detecting forged images across various generative models. Notably, even with just 1\% of the training data, our model exhibits minimal performance fluctuation, highlighting its ability to achieve competitive performance with a small amount of training data.

In summary, our contributions are:
\begin{itemize}[noitemsep, topsep=1pt]
    \itemsep 0em
    \item \textbf{Novel Feature-Guided CLIP-ViT for Forgery Image Detection.} We propose ForgeLens, a feature-guided CLIP-ViT framework for forgery image detection, ensuring strong generalization while maintaining high detection accuracy.
    \item \textbf{Effective Forgery Focus.} We introduce the lightweight Weight-Shared Guidance Module (WSGM) and FAFormer to enable the frozen pre-trained CLIP-ViT to focus on forgery-specific features during training. This effectively addresses the limitation of general-purpose image features extracted by CLIP-ViT, which contain excessive forgery-irrelevant information.
    \item \textbf{Effectiveness and Data-efficient.} We demonstrate the effectiveness of ForgeLens, which outperforms SOTA methods on UniversalFakeDetect dataset. Notably, it maintains superior performance even with extremely limited training data, highlighting strong data efficiency.
\end{itemize}

\section{Related Work and Background}

\subsection{Specialized Forgery Image Detection Methods}
Recent forgery image detection methods~\cite{rossler2019faceforensics++, liu2020global, chai2020makes, qian2020thinking, luo2021generalizing} use neural networks like Xception~\cite{chollet2017xception} and ResNet~\cite{he2016deep}, along with specialized modules, to extract forgery artifacts from the frequency and spatial domains of images to detect whether an image is generated.
Specifically, GLFF~\cite{ju2023glff} fuses multi-scale global features and local detail features extracted by a dual-branch network. Tan \textit{et al.}~\cite{tan2024frequency} propose FreqNet that integrates frequency domain learning into CNN classifiers to enhance their ability to capture frequency-based artifacts. NAFID~\cite{deng2023new} extracts non-local image features through a dedicated module. While effective at detecting forged images from generative models seen during training, these methods struggle with those from unseen models, showing limited generalization. Additionally, they require training from scratch and need large amounts of data to perform well. In contrast, our work achieves high generalization to unseen models, even with extremely limited training data.

\subsection{Preprocessing-based Detection Methods}
To address the challenge of generalizing to unseen models, existing methods have shifted towards data augmentation or transforming raw images into new representations through image preprocessing. Methods like CNN-Spot~\cite{wang2020cnn} employ diverse data augmentation techniques to enhance generalization to unseen testing data. Liu \textit{et al.}~\cite{liu2022detecting} identified a distinctive pattern in real images, referred to as Learned Noise Patterns, which exhibit stronger discrimination. FreGAN~\cite{jeong2022frepgan} integrates frequency-level perturbation maps to mitigate overfitting caused by unique artifacts in generated images, thereby enhancing detection accuracy across diverse test scenarios. BiHPF~\cite{jeong2022bihpf} amplifies the influence of frequency-level artifacts typically found in synthesized images from generative models using Bilateral High-Pass Filters. LGrad \cite{tan2023learning} transforms RGB images into their gradients through a transformation model, using these gradients as an image feature representation. NPR~\cite{tan2024rethinking} highlights the traces left by upsampling in forged images. However, while these methods improve generalization to some extent, they inevitably lose information from the original RGB image, which can affect detection accuracy. Additionally, they often require extra processing steps, making the detection pipeline more complex. In comparison, our end-to-end framework takes the original image as input and achieves better generalization effectively.

\subsection{Frozen Pretrained Model-Based Methods}
Another line of research~\cite{ojha2023towards, koutlis2024leveraging, khan2024clipping} uses a frozen pre-trained image encoder, such as CLIP-ViT~\cite{dosovitskiy2020image, radford2021learning}, to extract general-purpose image features, followed by a linear classifier for forgery detection. This approach helps prevent the model from overfitting to the unique forgery artifacts~\cite{corvi2023intriguing, ojha2023towards} in the training set and improves generalization to images generated by unseen generative models. In particular, FatFormer~\cite{liu2024forgery} proposes a forgery-aware adapter to aggregate frequency features and consider contrastive learning between image and text prompts used in CLIP. C2P-CLIP~\cite{tan2024c2p} injects category-related concepts into the image encoder via a category common prompt, improving detection performance. However, with network weights frozen during training, the features extracted by the image encoder are general-purpose and suitable for various downstream tasks. These features contain a large amount of forgery-irrelevant information, which limits detection performance. Unlike previous methods, our approach adopts an image encoder-only framework, which effectively enables the frozen image encoder to focus on forgery-specific features (e.g., face outlines, hair) while suppressing the influence of forgery-irrelevant information in the original general-purpose image features, thereby improving generalization and detection accuracy.
\begin{figure*}[t]
    \centering
    \includegraphics[width=1.0\linewidth]{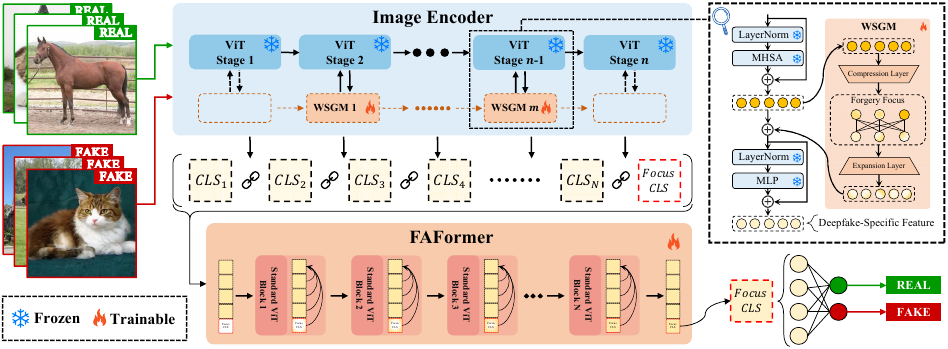}
    \caption{An illustration of the ForgeLens framework. The trainable WSGM is used to guide the frozen ViT in focusing on forgery-specific features; The CLS tokens (yellow square) extracted from each stage of the ViT are concatenated with a Focus CLS token and forwarded to FAFormer (red rectangle), which consists of standard ViT blocks and is used to focus on forgery features across multiple stages; The resulting Focus CLS token acts as the final representation for classification. The squares “$Focus\;CLS$” and “$CLS_{i}$” represent the additional Focus CLS token and the image CLS token output by the $i$-th stage of the frozen ViT, respectively. }
    \label{img3}
\end{figure*}

\section{Methods}
In this section, we introduce our \textit{ForgeLens} framework, designed for effective generalizable forgery image detection. Intuitively, ForgeLens consists of three steps: (1) It employs a weight-sharing guidance model (WSGM) to guide the frozen image encoder (CLIP-ViT) in focusing on forgery-specific features during feature extraction. (2) A forgery-aware feature integrator (FAFormer) then refines forgery-related information across multi-stage features. (3) Finally, the optimized image representation for forgery detection is fed into a linear classifier for the final prediction.

\subsection{Image Encoder}
Benefiting from CLIP’s powerful pre-trained weights and efficient feature extraction, we adopt CLIP-ViT~\cite{radford2021learning} as our frozen image encoder, following the strategy outlined in UniFD~\cite{ojha2023towards}. Given an RGB image $x \in \mathbb{R}^{H\times W\times 3}$, the patch embedding layer first splits it into $N$ non-overlapping patches and projects each patch into a $D$-dimensional feature space, forming a sequence of patch embeddings $E_{patch} \in \mathbb{R} ^{(N+1)\times D}$. Subsequently, the additional CLS token is prepended to the sequence, then the patch embeddings are processed through multiple Vision Transformer blocks, where self-attention mechanisms refine the representations. Finally, the output CLS token is used as the image representation, denoted as $f_{i} \in  \mathbb{R} ^{1\times D }$, at the $i$-th stage of CLIP-ViT. In this context, $H$ and $W$ denote the height and width of the input image, respectively, $N = \frac{HW}{P^{2}}$ is the total number of patches, $D$ represents the dimensionality of the projected features.

\subsection{Weight-Shared Guidance Module}
To address the issue of frozen pre-trained CLIP-ViT extracting general-purpose image features with excessive forgery-irrelevant information, we propose a lightweight, trainable Weight-Shared Guidance Module (WSGM) to guide the encoder toward forgery-specific features with minimal additional parameters.

In the standard Vision Transformer (ViT) pipeline, the Multi-Head Self-Attention (MHSA) module captures global features, while the Multi-Layer Perceptron (MLP) module refines and nonlinearly transforms these features. To guide the frozen ViT in focusing on forgery-specific features with minimal trainable parameters, we introduce the trainable WSGM, which operates across multiple ViT blocks and processes the features output by MHSA before they are fed into the MLP. WSGM optimizes the global features and emphasizes forgery-related information during training, ensuring that the MLP receives features suited for forgery detection. This allows the frozen encoder to progressively enhance its focus on forgery-specific features. The process can be formulated as follows: 
\begin{equation}\label{eqn-1} 
    \begin{aligned} 
        z_l &= \text{MHSA}(LN(x_{l-1})) + x_{l-1}, \quad l = 1, \dots, n \\
        z_l^{'} &= \text{WSGM}_k(z_l) + z_l, \quad k = 1, \dots, m \\
        x_l &= \text{MLP}(LN(z_l^{'})) + z_l^{'} 
    \end{aligned} 
\end{equation}
where $x_{l-1}$ and $x_{l}$ denote the input and output features of the $l$-th frozen ViT block, respectively. The WSGMs are shared across multiple ViT blocks, ensuring that each WSGM is responsible for exactly $\frac{n}{m}$ blocks, where $\frac{n}{m}$ is guaranteed to be an integer. Specifically, the $k$-th WSGM is applied to the MHSA output within its assigned block group before passing it to the MLP.

To further reduce the number of trainable parameters, the WSGM module adopts a bottleneck structure. It comprises compression and expansion layers, ReLU activations, and an additional linear layer specifically designed to guide ViT blocks in focusing on forgery-specific features, which can be formulated as:
\begin{equation}\label{eqn-2} 
    \begin{aligned}
        \text{WSGM}(z) = W_{com}\cdot \text{ReLU}(W_{mid}\cdot \text{ReLU}(W_{exp}\cdot z))
    \end{aligned}
\end{equation}
where $W_{com} \in \mathbb{R}^{d\times \hat{d} }$, $W_{exp} \in \mathbb{R}^{\hat{d}\times d}$ denotes compression and expansion projection matrices, $\hat{d}$ is the bottleneck middle dimension which satisfies $ \hat{d} \ll d $, $W_{mid} \in \mathbb{R}^{\hat{d}\times \hat{d}}$ denote the additional middle linear layer.
 
The WSGM module effectively guides the frozen pre-trained CLIP-ViT to focus on forgery-specific features, addressing the accuracy degradation caused by the excessive forgery-irrelevant information in the general-purpose features extracted by the frozen network.

\subsection{FAFormer}
After guiding the extraction of forgery-specific features, we further consider the importance of shallow image features in forgery detection~\cite{bayar2018constrained, corvi2023detection, koutlis2024leveraging}. Therefore, we propose FAFormer, a forgery-aware feature integrator that refines forgery-related information across multi-stage features, including both low-level and high-level representations.

Building on the finding that the classical Transformer \cite{vaswani2017attention} utilizes the CLS token to capture a comprehensive representation of the input sequence, this mechanism enables the model to aggregate diverse feature representations into a unified embedding. As a global representation, the CLS token integrates information from all input tokens through self-attention, effectively refining the most relevant information from the input sequence. Furthermore, we observe that the output CLS tokens $CLS_{i} \in \mathbb{R}^{D}$ from each stage of the frozen ViT model inherently satisfy the input requirements of a Transformer, eliminating the need for additional embedding transformations. Thus, we concatenate the CLS tokens from each stage and introduce a new Focus CLS token $CLS_{focus} \in \mathbb{R}^{D}$, which acts as a global representation, refining information from all stages. This composite representation is expressed as:
\begin{equation}\label{eqn-3} 
    c_{0} = [CLS_{focus}; CLS_{1}; CLS_{2}; \dots ; CLS_{N}]
\end{equation}

Based on this, FAFormer is designed to refine forgery-specific information across multi-stage features, encompassing both low-level and high-level image features. The module follows the standard Vision Transformer (ViT) block architecture \cite{dosovitskiy2020image}, and its process can be formulated as:
\begin{equation}\label{eqn-4}
	\begin{aligned}
		c_{l}^{'} &= \text{MHSA}(LN(c_{l-1})) + c_{l-1}\\
		c_{l} &= \text{MLP}(LN(c_{l}^{'})) + c_{l}^{'}
	\end{aligned}
\end{equation}
where $c_{l}^{'}$, $c_{l}$ represent the output features of the MHSA and MLP modules for standard ViT block $l$, respectively.

The FAFormer uses the Transformer's self-attention mechanism to effectively integrate multi-stage features extracted from the frozen CLIP-ViT, combining fine-grained low-level details with abstract high-level semantics. Low-level features capture subtle textures and artifacts, while high-level features encode more semantic and structural information. By refining these diverse features, FAFormer enhances the representation of forgery-specific cues, ensuring that both fine-grained and abstract information contributes to forgery detection. 
\begin{table*}[ht]
    \centering
    \resizebox{1.0\textwidth}{!}{
    \begin{tabular}{l c c c c c c c c c c c c c c c c c c c c c c}
    \toprule
         \multirow{3}{*}{Method} & 
         \multicolumn{6}{c}{Generative Adversarial Networks} & 
         \multirow{3}{*}{\makecell{Deep\\fakes}} & 
         \multicolumn{2}{c}{Low level vision} & 
         \multicolumn{2}{c}{Perceptual loss} & 
         \multirow{3}{*}{Guided} & 
         \multicolumn{3}{c}{LDM} & 
         \multicolumn{3}{c}{Glide} & 
         \multirow{3}{*}{DALL-E} & 
         \multirow{3}{*}{\makecell{Avg.\\Acc}} \\
         
         \cmidrule(lr){2-7}
         \cmidrule(lr){9-10}
         \cmidrule(lr){11-12}
         \cmidrule(lr){14-16}
         \cmidrule(lr){17-19}

         & \makecell{Pro\\GAN} & \makecell{Cycle\\GAN} & \makecell{Big\\GAN} & \makecell{Style\\GAN} & \makecell{Gau\\GAN} & \makecell{Star\\GAN} & 
         & SITD & SAN & CRN & IMLE & 
         & \makecell{200\\steps} & \makecell{200\\w/CFG} & \makecell{100\\steps} & 
         \makecell{100\\27} & \makecell{50\\27} & \makecell{100\\10} & 
         & \\
        \midrule
        \textit{\textbf{Specialized}} &&&&&&&&&&&&&&&&&&&& \\
        Patchfor~\cite{chai2020makes} & 75.03 & 68.97 & 68.47 & 79.16 & 64.23 & 63.94 & 75.54 & 75.14 & 75.28 & 72.33 & 55.30 & 67.41 & 76.50 & 76.10 & 75.77 & 74.81 & 73.28 & 68.52 & 67.91 & 71.24 \\
        F3Net~\cite{qian2020thinking} & 99.38 & 76.38 & 65.33 & 92.56 & 58.10 & 100.0 & 63.48 & 54.17 & 47.26 & 51.47 & 51.47 & 96.20 & 68.15 & 75.35 & 68.80 & 81.65 & 83.25 & 83.05 & 66.30 & 71.33 \\
        FreqNet~\cite{tan2024frequency} & 97.90 & 95.84 & 90.45 & 97.55 & 90.24 & 93.41 & 97.40 & 88.92 & 59.04 & 71.92 & 67.35 & 86.70 & 84.55 & 99.58 & 65.56 & 85.69 & 97.40 & 88.15 & 59.06 & 85.09 \\
        \midrule
        \textit{\textbf{Preprocessing-based}} &&&&&&&&&&&&&&&&&&&& \\
        CNN-Spot~\cite{wang2020cnn} & 99.99 & 85.20 & 70.20 & 85.70 & 78.95 & 91.70 & 53.47 & 66.67 & 48.69 & 86.31 & 86.26 & 60.07 & 54.03 & 54.96 & 54.14 & 60.78 & 63.80 & 65.66 & 55.58 & 69.58 \\
        LGrad~\cite{tan2023learning} & 99.84 & 85.39 & 82.88 & 94.83 & 72.45 & 99.62 & 58.00 & 62.50 & 50.00 & 50.74 & 50.78 & 77.50 & 94.20 & 95.85 & 94.80 & 87.40 & 90.70 & 89.55 & 88.35 & 80.28 \\
        NPR~\cite{tan2024rethinking} & 99.84 & 95.00 & 87.55 & 96.23 & 86.57 & 99.75 & 76.89 & 66.94 & 98.63 & 50.00 & 50.00 & 84.55 & 97.65 & 98.00 & 98.20 & 96.25 & 97.15 & 97.35 & 87.15 & 87.56 \\
        \midrule
        \textit{\textbf{Frozen Model-based}} &&&&&&&&&&&&&&&&&&&& \\
        UniFD~\cite{ojha2023towards} & 100.0 & 98.50 & 94.50 & 82.00 & 99.50 & 97.00 & 66.60 & 63.00 & 57.50 & 59.50 & 72.00 & 70.03 & 94.19 & 73.76 & 94.36 & 79.07 & 79.85 & 78.14 & 86.78 & 81.38 \\
        FatFormer~\cite{liu2024forgery} & 99.89 & 99.32 & 99.50 & 97.15 & 99.41 & 99.75 & 93.23 & 81.11 & 68.04 & 69.45 & 69.45 & 76.00 & 98.60 & 94.90 & 98.65 & 94.35 & 94.65 & 94.20 & 98.75 & 90.86 \\
        RINE~\cite{koutlis2024leveraging} & 100.0 & 99.30 & 99.60 & 88.90 & 99.80 & 99.50 & 80.60 & 90.60 & 68.30 & 89.20 & 90.60 & 76.10 & 98.30 & 88.20 & 98.60 & 88.90 & 92.60 & 90.70 & 95.00 & 91.31\\
        C2P-CLIP~\cite{tan2024c2p} & 99.98 & 97.31 & 99.12 & 96.44 & 99.17 & 99.60 & 93.77 & 95.56 & 64.38 & 93.29 & 93.29 & 69.10 & 99.25 & 97.25 & 99.30 & 95.25 & 95.25 & 96.10 & 98.55 & \underline{93.79} \\
        \midrule
        \textit{\textbf{Ours}} &&&&&&&&&&&&&&&&&&&& \\
        FreLens & 99.95 & 99.24 & 97.67 & 96.64 & 98.84 & 95.24 & 88.97 & 85.83 & 93.75 & 97.23 & 97.55 & 73.34 & 98.72 & 96.98 & 98.86 & 96.07 & 96.17 & 95.43 & 98.29 & \textbf{94.99} \\
    \bottomrule
\end{tabular}}
\caption{Accuracy (Acc) results of forgery detection methods on UniversalFakeDetect, covering both GANs and diffusion models. Methods are categorized into \textit{\textbf{Specialized}} methods $|$ \textit{\textbf{Preprocessing-based}} methods $|$ \textit{\textbf{Frozen Model-based}} methods $|$ \textit{\textbf{Ours}}. \textbf{Bold} and \underline{underline} represent the best and second-best performance, respectively.}
\label{tab:main_result1}
\end{table*}

\begin{table*}[ht]
    \centering
    \resizebox{1.0\textwidth}{!}{
    \begin{tabular}{l c c c c c c c c c c c c c c c c c c c c c c}
    \toprule
         \multirow{3}{*}{Method} & 
         \multicolumn{6}{c}{Generative Adversarial Networks} & 
         \multirow{3}{*}{\makecell{Deep\\fakes}} & 
         \multicolumn{2}{c}{Low level vision} & 
         \multicolumn{2}{c}{Perceptual loss} & 
         \multirow{3}{*}{Guided} & 
         \multicolumn{3}{c}{LDM} & 
         \multicolumn{3}{c}{Glide} & 
         \multirow{3}{*}{DALL-E} & 
         \multirow{3}{*}{\makecell{Avg.\\Acc}} \\
         
         \cmidrule(lr){2-7}
         \cmidrule(lr){9-10}
         \cmidrule(lr){11-12}
         \cmidrule(lr){14-16}
         \cmidrule(lr){17-19}

         & \makecell{Pro\\GAN} & \makecell{Cycle\\GAN} & \makecell{Big\\GAN} & \makecell{Style\\GAN} & \makecell{Gau\\GAN} & \makecell{Star\\GAN} & 
         & SITD & SAN & CRN & IMLE & 
         & \makecell{200\\steps} & \makecell{200\\w/CFG} & \makecell{100\\steps} & 
         \makecell{100\\27} & \makecell{50\\27} & \makecell{100\\10} & 
         & \\
        \midrule
        \textit{\textbf{Specialized}} &&&&&&&&&&&&&&&&&&&& \\
        Patchfor & 80.88 & 72.84 & 71.66 & 85.75 & 65.99 & 69.25 & 76.55 & 76.19 & 76.34 & 74.52 & 68.52 & 75.03 & 87.10 & 86.72 & 86.40 & 85.37 & 83.73 & 78.38 & 75.67 & 77.73 \\
        F3Net & 99.96 & 84.32 & 69.90 & 99.72 & 56.71 & 100.0 & 78.82 & 52.89 & 46.70 & 63.39 & 64.37 & 70.53 & 73.76 & 81.66 & 74.62 & 89.81 & 91.04 & 90.86 & 71.84 & 76.89 \\
        FreqNet & 99.92 & 99.63 & 96.05 & 99.89 & 99.71 & 98.63 & 99.92 & 94.42 & 74.59 & 80.10 & 75.70 & 96.27 & 96.06 & 100.0 & 62.34 & 99.80 & 99.78 & 96.39 & 77.78 & 91.95 \\
        \midrule
        \textit{\textbf{Preprocessing-based}} &&&&&&&&&&&&&&&&&&&& \\
        CNN-Spot & 100.0 & 93.47 & 84.50 & 99.54 & 89.49 & 98.15 & 89.02 & 73.75 & 59.47 & 98.24 & 98.40 & 73.72 & 70.62 & 71.00 & 70.54 & 80.65 & 84.91 & 82.07 & 70.59 & 83.58 \\
        LGrad & 100.0 & 93.98 & 90.69 & 99.86 & 79.36 & 99.98 & 67.91 & 59.42 & 51.42 & 63.52 & 69.61 & 87.06 & 99.03 & 99.16 & 99.18 & 93.23 & 95.10 & 94.93 & 97.23 & 86.35 \\
        NPR & 100.0 & 99.53 & 94.53 & 99.94 & 88.82 & 100.0 & 84.41 & 97.95 & 99.99 & 50.16 & 50.16 & 98.26 & 99.92 & 99.91 & 99.92 & 99.87 & 99.89 & 99.92 & 99.26 & 92.76 \\
        \midrule
        \textit{\textbf{Frozen Model-based}} &&&&&&&&&&&&&&&&&&&& \\
        UniFD & 100.0 & 98.13 & 94.46 & 86.66 & 99.25 & 99.53 & 91.67 & 78.54 & 67.54 & 83.12 & 91.06 & 79.24 & 95.81 & 79.77 & 95.93 & 93.93 & 95.12 & 94.59 & 88.45 & 90.14 \\
        FatFormer & 100.0 & 100.0 & 99.98 & 99.75 & 100.0 & 100.0 & 97.99 & 97.94 & 81.21 & 99.84 & 99.93 & 91.99 & 99.81 & 99.09 & 99.87 & 99.13 & 99.41 & 99.20 & 99.82 & 98.16 \\
        RINE & 100.0 & 100.0 & 99.90 & 99.40 & 100.0 & 100.0 & 97.90 & 97.20 & 94.90 & 97.30 & 99.70 & 96.40 & 99.80 & 98.30 & 99.90 & 98.80 & 99.30 & 98.90 & 99.30 & \underline{98.78}\\
        C2P-CLIP & 100.0 & 100.0 & 99.96 & 99.50 & 100.0 & 100.0 & 98.59 &98.92 & 84.56 & 99.86 & 99.95 & 94.13 & 99.99 & 99.83 & 99.98 & 99.72 & 99.79 & 99.83 & 99.91 & 98.66 \\
        \midrule
        \textit{\textbf{Ours}} &&&&&&&&&&&&&&&&&&&& \\
        FreLens & 100.0 & 100.0 & 99.83 & 99.82 & 99.98 & 100.0 & 95.44 & 94.20 & 98.69 & 99.94 & 99.99 & 92.92 & 99.92 & 99.49 & 99.87 & 99.10 & 99.45 & 99.33 & 99.80 & \textbf{98.83} \\
    \bottomrule
\end{tabular}}
\caption{Average Precision (AP) results of forgery detection methods on UniversalFakeDetect.}
\label{tab:main_result2}
\end{table*}

\section{Experimental Setup}
\subsection{Datasets}
\noindent{\textbf{Training Dataset.}}
To ensure a fair and consistent comparison, we follow existing methods by using only ProGAN-generated images for training.  Specifically, we use the ForenSynths \cite{wang2020cnn} training and validation sets in line with previous methods~\cite{ojha2023towards, liu2024forgery, tan2024c2p}. The ForenSynths training set comprises 20 categories of real images sourced from the LSUN dataset~\cite{yu2015lsun}, along with their corresponding fake images synthesized by ProGAN~\cite{karras2017progressive}. Following prior work, we adopt a 4-class subset (car, cat, chair, horse) for training. To further highlight the data efficiency of our method, which requires only a limited amount of training data, we partition the training set into subsets containing 1\%, 4\%, 20\%, 50\%, and 100\% of the data. More details about the training set are provided in Appendix \ref{appendix:training_data}.

\noindent{\textbf{Evaluation Datasets.}}
To evaluate the effectiveness and generalization ability of our approach, we adopt the widely used dataset, UniversalFakeDetect\cite{ojha2023towards}. Specifically, the UniversalFakeDetect test set contains 19 subsets across various generative models, including ProGAN~\cite{karras2017progressive}, StyleGAN~\cite{karras2019style}, BigGAN~\cite{brock2018large}, CycleGAN~\cite{zhu2017unpaired}, StarGAN~\cite{choi2018stargan}, GauGAN~\cite{park2019semantic}, Deepfake~\cite{rossler2019faceforensics++}, CRN~\cite{chen2017photographic}, IMLE~\cite{li2019diverse}, SAN~\cite{dai2019second}, SITD~\cite{chen2018learning}, Guided diffusion model~\cite{dhariwal2021diffusion}, LDM~\cite{rombach2022high}, Glide~\cite{nichol2021glide}, DALLE~\cite{ramesh2021zero}. Detailed information is provided in Appendix~\ref{appendix:evaluation_data}.

\subsection{Baselines}
To comprehensively evaluate our method, we compare it against various forgery detection approaches, categorized into specialized detection methods, preprocessing-based methods, and frozen model-based methods. Specialized detection methods include Patchfor~\cite{chai2020makes}, F3Net~\cite{qian2020thinking} and FreqNet~\cite{tan2024frequency}. For preprocess-based detection methods, we include CNN-Spot~\cite{wang2020cnn}, LGrad~\cite{tan2023learning}, and NPR~\cite{tan2024rethinking}. For frozen model-based detection methods, we include UniFD~\cite{ojha2023towards}, FatFormer~\cite{liu2024forgery}, RINE~\cite{koutlis2024leveraging} and C2P-CLIP~\cite{tan2024c2p}.

\subsection{Evaluation metric}
Following previous works~\cite{ojha2023towards, koutlis2024leveraging, liu2024forgery}, we use accuracy (Acc) and Average Precision (AP) as the primary metrics to evaluate the effectiveness of our method. To comprehensively assess model generalization across various GAN and diffusion model subsets, we report the average Acc and AP.

\subsection{Implementation Details}
We employ a two-stage training strategy to ensure stable training. In the first stage, we freeze the CLIP-ViT backbone and train only the WSGM. In the second stage, we keep all previous networks frozen and integrate FAFormer into the backend, training it independently.
Input images are randomly cropped to a resolution of 224 × 224. During training, we apply horizontal flipping without additional augmentations. For testing, only center cropping is applied. We use the Adam optimizer \cite{kingma2014adam} with beta parameters set to (0.9, 0.999) and employ binary cross-entropy loss. All experiments were conducted using the PyTorch framework \cite{paszke2019pytorch} on an Nvidia GeForce RTX 3090 GPU. Additional details are provided in Appendix \ref{appendix: hyperparameter}.

\section{Results and Analysis}
\subsection{Comparison with competing methods} 
Table \ref{tab:main_result1} and Table \ref{tab:main_result2} present the Acc and AP of our proposed method, ForgeLens, compared to existing forgery detection methods on UniversalFakeDetect~\cite{ojha2023towards}. The results show that our proposed ForgeLens outperforms current methods, achieving strong generalization, reaching an average Acc of 94.99\% and and average AP of 98.83\% across all subsets. 

Specifically, among specialized methods, the recent approach FreqNet achieves the best performance on GAN subsets related to the training set, as it was trained on ProGAN. However, it shows limitations when applied to diffusion model subsets. In contrast, our method exhibits high generalization across all subsets, achieving 9.9\% improvement in average Acc and 6.88\% improvement in average AP over FreqNet. For preprocessing-based methods, our approach outperforms them, with an average Acc 7.43\% and an average AP 6.07\% higher than the best method NPR, achieving a comprehensive improvement in accuracy across various subsets. For frozen model-based methods, compared to the base model UniFD, which uses the original frozen CLIP-ViT to extract image features for classification, our method improves the average Acc by 13.61\% and the average AP by 8.69\%. Additionally, our approach outperforms RINE, an advanced method based on UniFD. This demonstrates that our approach effectively addresses the issue of general-purpose image features extracted from the frozen pre-trained encoder, which are unsuitable for forgery detection due to containing excessive forgery-irrelevant information. Notably, compared to FatFormer and the latest state-of-the-art method, C2P-CLIP, both of which use both text and image encoders, thereby increasing model complexity, our method achieves superior performance with a simpler architecture, improving 3.68\% and 1.2\% average Acc over them, respectively. 

\begin{table}[t]
\centering
\newcommand{\cmark}{\ding{51}}
\newcommand{\xmark}{\ding{55}}
\resizebox{0.45\textwidth}{!}{
\begin{tabular}{c c c c}
    \toprule
    w/ WSGM & W/ FAFormer & Avg.Acc.(\%) & Avg.AP.(\%) \\
    \midrule
    \xmark & \xmark  & 81.38 & 90.14\\
    \xmark & \cmark  & 87.89 & 92.26\\
    \cmark & \xmark  & 94.52 & 98.12\\
    \cmark & \cmark  & \textbf{94.99} & \textbf{98.83}\\
    \bottomrule
\end{tabular}}
\caption{Average accuracy and average precision (Acc/AP) in the Ablation Study of WSGM and FAFormer on UniversalFakeDetect. Results without each module are denoted as 'w/o'. The top-performing results are highlighted in bold.}
\label{tab: ablation1}
\end{table}

\subsection{Ablation Study}
\noindent{\textbf{Effectiveness Analysis of Forgery Focus.}}
We conducted an ablation experiment to validate the effectiveness of WSGM and FAFormer on UniversalFakeDetect and GenImage. The results are detailed in Table \ref{tab: ablation1}. Compared to the base model, the introduction of WSGM and FAFormer leads to significant improvements, particularly with WSGM, which improves Avg.ACC by 13.14\% and Avg.AP by 7.98\%. FAFormer further refines forgery features across various stages, leading to additional performance gains.

\begin{figure}[t]
    \centering
    \includegraphics[width=0.95\linewidth]{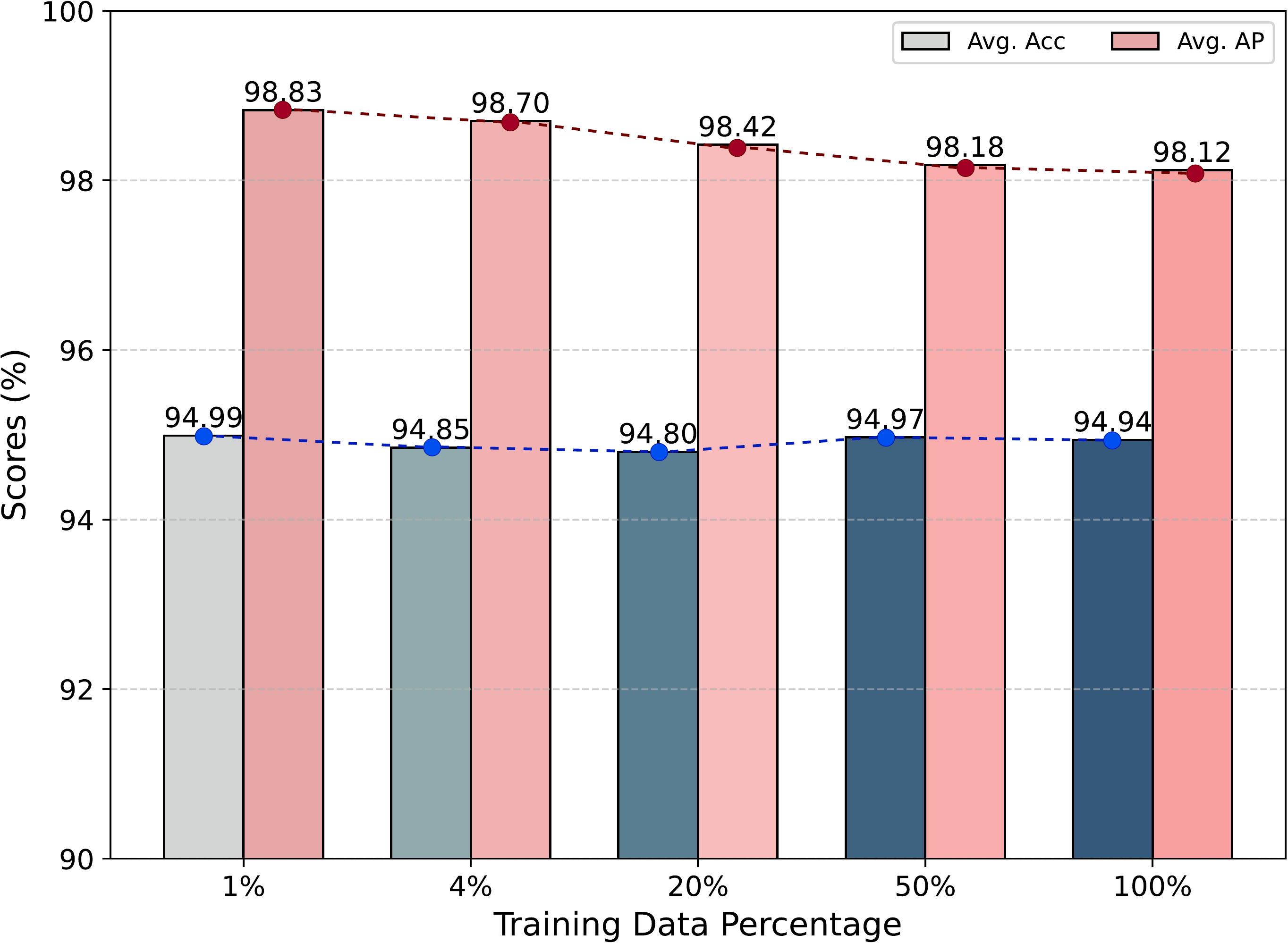}
    \caption{Average Acc and Average AP comparison across different training data sizes on UniversalFakeDetect. The bar colors deepen as the training data percentage increases.}
    \label{fig: datasize}
\end{figure}

\noindent{\textbf{Impact of Different Training Data Size.}}
\label{sec: data_size}
To demonstrate the data efficiency of our approach, we evaluate the impact of different training data sizes (1\%, 4\%, 20\%, 50\%, and 100\%) on model performance, as shown in Figure \ref{fig: datasize}. The results show that our method performs well and remains stable across all settings. Interestingly, as the amount of training data increases, the Avg.AP value tends to decrease slightly. Notably, even with only 1\% of the training data, it achieves optimal performance comparable to that with 100\% of the data. This further validates the effectiveness of our approach, highlighting its ability to adapt and counter emerging forgery techniques quickly.

\noindent{\textbf{Comparative Analysis of WSGM and Fine-Tuning Methods.}}
To further validate the effectiveness of our proposed WSGM, we compare it with previous fine-tuning methods, Adapter~\cite{chen2022adaptformer} and LoRA~\cite{hu2022lora}. We conduct experiments by applying each method directly to the base model CLIP-ViT. The results in Table \ref{tab: ablation2} show that our method significantly outperforms both Adapter and LoRA. Unlike fine-tuning methods that merely adjust trainable parameters to adapt to a specific task, WSGM employs a guidance mechanism that guides the frozen CLIP-ViT to focus on forgery-specific features.

\begin{table}[t]
\centering
\newcommand{\cmark}{\ding{51}}
\newcommand{\xmark}{\ding{55}}
\resizebox{0.45\textwidth}{!}{
\begin{tabular}{c c c c}
    \toprule
    Method & Backbone & Avg.Acc.(\%) & Avg.AP.(\%) \\
    \midrule
    \xmark & CLIP-ViT & 81.38 & 90.14 \\
    \midrule
    Adapter & CLIP-ViT & 85.49 (4.11$\uparrow$) & 94.29 (4.15$\uparrow$)\\
    LoRA & CLIP-ViT & 86.96 (5.58$\uparrow$) & 96.41 (6.27$\uparrow$) \\
    WSGM (Ours) & CLIP-ViT & \textbf{94.52 (12.86$\uparrow$)} & \textbf{98.12 (7.98$\uparrow$)} \\
    \bottomrule
\end{tabular}}
\caption{Average Accuracy (Avg.Acc) and Average Precision (Avg.Ap) evaluated on UniversalFakeDetect. We compare our proposed WSGM with the previous fine-tuning method Adapter and LoRA. The top-performing results are highlighted in bold.}
\label{tab: ablation2}
\end{table}

\begin{figure*}[t]
\centering
    \includegraphics[width=0.95\linewidth]{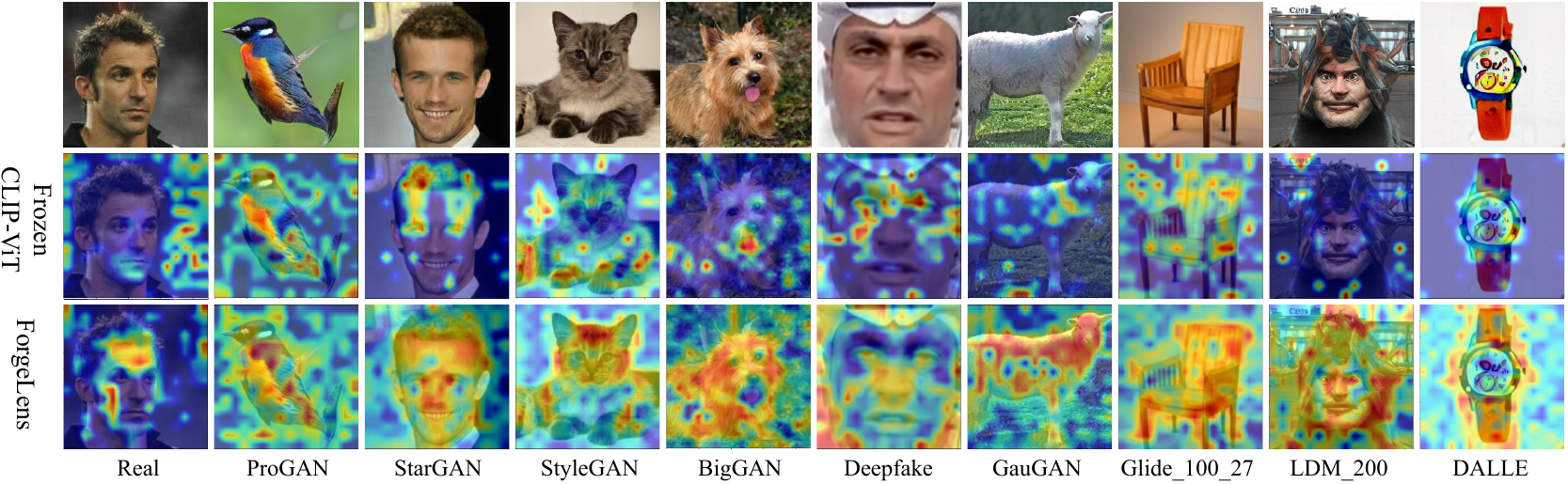}
    \caption{Class Activation Map (CAM) \cite{zhou2016learning} visualization of features extracted by the frozen CLIP-ViT and our proposed ForgeLens.}
\label{img4}
\end{figure*} 

\subsection{Robustness Evaluation}
Considering that images transmitted on the internet are often subjected to postprocessing in real-world scenarios, we evaluated the robustness of our method under various perturbations by independently applying Gaussian noise, Gaussian blur, JPEG compression, and random cropping to both the training and test sets. Further details are provided in Appendix~\ref{appendix:perburbation}. Additionally, to simulate real-world conditions, we sequentially applied all four perturbations simultaneously, referred to as the "combined" perturbation. The results are presented in Table \ref{tab: robust}. 

\begin{table}[ht]
\centering
\newcommand{\cmark}{\ding{51}}
\newcommand{\xmark}{\ding{55}}
\resizebox{0.45\textwidth}{!}{
\begin{tabular}{c c c c c c}
    \toprule
    w/Blur & w/Cropping & w/JPEG & w/Noise & Avg.Acc.(\%) & Avg.AP.(\%) \\
    \midrule
    \cmark & \xmark & \xmark & \xmark & 81.98 (13.01$\downarrow$) & 92.10 (6.73$\downarrow$) \\
    \xmark & \cmark & \xmark & \xmark & 91.17 (3.28$\downarrow$) & 96.97 (1.86$\downarrow$)\\
    \xmark & \xmark & \cmark & \xmark & 91.62 (3.37$\downarrow$) & 97.74 (1.09$\downarrow$)\\
    \xmark & \xmark & \xmark & \cmark & 82.64 (12.35$\downarrow$)& 93.41 (5.42$\downarrow$)\\
    \cmark & \cmark & \cmark & \cmark & 86.54 (8.45$\downarrow$) & 94.87 (3.96$\downarrow$)\\
    \bottomrule
\end{tabular}}
\caption{Average Accuracy (Avg.Acc) and Average Precision (Avg.Ap) evaluated on UniversalFakeDetect under various perturbations.}
\label{tab: robust}
\end{table}

The experimental results show that our method is minimally affected by random cropping and JPEG compression, with only a 3\% average accuracy decrease. However, Gaussian blur and noise have a more significant impact, causing an accuracy drop of 13.01\% and 12.35\%, respectively. This is because these perturbations impair image quality and weaken the forged artifacts in the image, therefore making it difficult for our method to effectively focus on the forge-specific features. Moreover, when considering all perturbations simultaneously, our method still demonstrates strong robustness, achieving an average accuracy of 86.54\% and an average AP of 94.87\%, outperforming the base model. This highlights the effectiveness of our method in real-world scenarios.

\subsection{Class Activation Map Visualization}
To further illustrate the impact of focusing on forgery-specific features and improve explainability, we conduct a class activation map (CAM) analysis on the features extracted by the frozen base model CLIP-ViT and our proposed ForgeLens, from both real images and forgery images generated by various generative models, as shown in Figure \ref{img4}. The visualization results confirm that the frozen pre-trained CLIP-ViT model mainly captures general-purpose image features, making it broadly suitable for various downstream tasks but ineffective in specifically detecting forgeries. As a result, it often fails to accurately identify forgery artifacts, such as facial outlines, hair textures, etc. This limitation is evident in the CAMs (second row), which reveal a dispersed attention pattern and an inability to localize forged regions effectively. In contrast, the features extracted by ForgeLens (third row) focus on specific forged regions in deepfake images, demonstrating a more focused and targeted response. For instance, when detecting DeepFake images, ForgerLens significantly highlights the regions with face swap artifacts. These results reaffirm that ForgeLens effectively overcomes the limitations of the general image features extracted by the frozen CLIP-ViT model, providing a more specialized feature representation that enhances detection accuracy and generalization.
\section{Conclusion}
In this work, we introduced ForgeLens, a data-efficient and feature-guided framework for generalizable forgery detection. By integrating the lightweight Weight-Shared Guidance Module (WSGM) to guide forgery focus and FAFormer to refine forgery-aware features across multiple stages, ForgeLens effectively addresses the limitations of previous frozen network-based methods, reducing forgery-irrelevant information while improving feature discrimination. Extensive experiments on 19 generative models, including both GANs and diffusion models, demonstrate that ForgeLens achieves state-of-the-art performance while requiring only minimal training data.

\noindent{\textbf{Limitations and Future Works.}}
Admittedly, one limitation of our approach is the use of a two-stage training strategy to ensure stability, which slightly increases overall training complexity. Additionally, in the second stage, FAFormer further refines forgery-specific features across multiple ViT stages, making the model more sensitive to hyperparameter selection and requiring careful tuning. Future work will focus on addressing these challenges to improve both the efficiency and robustness of our framework.

\section*{Acknowledgments}
Yakun Niu is supported by the National Natural Science Foundation of China (Grant 62202141). Lei Zhang is supported by the Key R\&D Projects in Henan Province (Grant 241111212800), the Scientific and Technological Key Project in Henan Province (Grant 232102240020), and the China University Research Innovation Fund (Grant 2023ZB014). This work is also supported by the Henan Province University Science and Technology Innovation Team Support Plan (Grant 24IRTSTHN021).

{
    \small
    \bibliographystyle{ieeenat_fullname}
    \bibliography{main}
}

\clearpage
\appendix
\label{sec:appendix}

\section{Dataset Details}
\subsection{Training Dataset}
\label{appendix:training_data}
We use the ForenSynths \cite{wang2020cnn} training set for our experiments. The details of the dataset are provided in Table \ref{tab: traindata}. For different data volume divisions, we randomly extract approximately 1\%, 4\%, 20\%, and 50\% of the data from the original training set.
\begin{table}[h]
    \centering
    \resizebox{0.4\textwidth}{!}{
    \begin{tabular}{lccc}
        \toprule
        Volume & Total Size & Classes & Neg\% \\
        \midrule
        1\%   & 1,600  & car, cat, chair, horse & 50\% \\
        4\%   & 6,400  & car, cat, chair, horse & 50\% \\
        20\%  & 28,800 & car, cat, chair, horse & 50\% \\
        50\%  & 72,000 & car, cat, chair, horse & 50\% \\
        100\% & 144,024 & car, cat, chair, horse & 50\% \\
        \bottomrule
    \end{tabular}}
    \caption{Statistics of Training Dataset. We report the number of images in each data split, the class distribution (all classes share the total data volume), and the proportion of negative samples.}
    \label{tab: traindata}
\end{table}

\subsection{Evaluation Datasets}
\label{appendix:evaluation_data}
\noindent\textbf{UniversalFakeDetect~\cite{ojha2023towards}.} The dataset is a large-scale benchmark designed to evaluate the generalization capability of forgery detection models across different generative techniques. It consists of image subsets generated by 19 different generative models, including both GAN-based and diffusion-based methods. Each subset contains both real and fake images, with some subsets further divided based on image classes. The details are shown in Table~\ref{tab: evaluateddata}.

\begin{table}[h]
    \centering
    \resizebox{0.4\textwidth}{!}{
    \begin{tabular}{lccc}
        \toprule
        Generative Models & Size & Class Count & Neg\% \\
        \midrule
        ProGAN   & 8,000  & 20 & 50\% \\
        CycleGAN & 2,642  & 6 & 50\% \\
        BigGAN  & 4,000 & N/A & 50\% \\
        StyleGAN  & 11,982 & 3 & 50\% \\
        GauGAN & 10,000 & N/A & 50\% \\
        StarGAN & 3,998 & N/A & 50\% \\
        Deepfakes & 5,405 & N/A & 49.9\% \\
        SITD & 360 & N/A & 50\% \\
        SAN & 438 & N/A & 50\% \\
        CRN & 12,764 & N/A & 50\% \\
        IMLE & 12,764 & N/A & 50\% \\
        Guided & 2,000 & N/A & 50\% \\
        LDM 200 steps & 3,000 & N/A & 33.3\% \\
        LDM 200 w/CFG & 3,000 & N/A & 33.3\% \\
        LDM 100 steps & 3,000 & N/A & 33.3\% \\
        Glide-100-27 & 3,000 & N/A & 33.3\% \\
        Glide-50-27 & 3,000 & N/A & 33.3\% \\
        Glide-100-10 & 3,000 & N/A & 33.3\% \\
        DALL-E & 3,000 & N/A & 33.3\% \\
        \bottomrule
    \end{tabular}}
    \caption{Statistics of the UniversalFakeDetect Dataset. We report the size of each subset, the number of classes (N/A indicates no class split), and the proportion of negative samples.}
    \label{tab: evaluateddata}
\end{table}

\section{Perturbation Details}
\label{appendix:perburbation}

\noindent{\textbf{Noise:}} Gaussian noise is added to the input image, with the variance randomly sampled from a uniform distribution in the range [5.0, 20.0]. This variance determines the noise intensity, introducing random pixel variations while preserving the original image dimensions.

\noindent{\textbf{Blurring:}} A Gaussian blur is applied using a kernel size randomly selected from {3, 5, 7, 9}. Larger kernels result in stronger blurring effects.

\noindent{\textbf{Compression:}} JPEG compression is introduced by selecting a random quality factor between 10 and 75. The image is then encoded in JPEG format with the chosen quality, inducing lossy compression artifacts.

\noindent{\textbf{Cropping:}} A random crop is performed by selecting a cropping ratio between 5\% and 20\% for both x and y dimensions. The cropped region is then resized back to the original dimensions using bicubic interpolation.

\begin{table}[t]
\centering
\resizebox{0.35\textwidth}{!}{
\begin{tabular}{cccc}
    \toprule
    \makecell{WSGM\\count} & \makecell{WSGM\\reduction\_factor} & Avg.Acc.(\%) & Avg.AP.(\%) \\
    \midrule
    4  & 2  & 94.42 & 98.81 \\
    4  & 4  & 91.12 & 98.31 \\
    4  & 8  & 92.07 & 98.15 \\
    \midrule
    8  & 2  & 93.61 & 98.55 \\
    8  & 4  & 94.92 & 98.78 \\
    8  & 8  & 91.99 & 98.31 \\
    \midrule
    12  & 2  & 93.24 & 98.23 \\
    \textbf{12}  & \textbf{4}  & \textbf{94.99} & \textbf{98.83} \\
    12  & 8  & 93.86 & 98.35 \\
    \midrule
    24  & 2  & 93.11 & 98.16 \\
    24  & 4  & 94.74 & 98.79 \\
    24  & 8  & 90.12 & 97.37 \\
    \bottomrule
\end{tabular}}
\caption{Ablation Experiment Results on WSGM Count and Reduction Factor.}
\label{tab: wsgm_ablation}
\end{table}

\begin{table}[b]
\centering
\resizebox{0.35\textwidth}{!}{
\begin{tabular}{cccc}
    \toprule
    FAFormer Layers & Avg.Acc.(\%) & Avg.AP.(\%) \\
    \midrule
    1   & 94.63 (0.11$\uparrow$) & 98.72 (0.60$\uparrow$) \\
    2   & 94.99 (0.47$\uparrow$) & 98.83 (0.71$\uparrow$) \\
    4   & 94.91 (0.39$\uparrow$) & 99.00 (0.88$\uparrow$) \\
    6   & 94.78 (0.26$\uparrow$) & 98.81 (0.69$\uparrow$)\\
    8   & 94.94 (0.42$\uparrow$) & 98.86 (0.74$\uparrow$)\\
    \bottomrule
\end{tabular}}
\caption{Ablation Experiment Results on FAFormer Layers.}
\label{tab: faformer_ablation}
\end{table}

\section{Additional Experimental results}
\subsection{Ablation Studies}

\noindent\textbf{Impact of WSGM Block Count and Bottleneck Dimension.} To evaluate the impact of different WSGM counts and their bottleneck dimension, we conduct experiments to analyze their effects on detection performance. The results are presented in Table~\ref{tab: wsgm_ablation}. The experimental results show that different WSGM counts and their reduction factors have some impact on ACC, but the effect on AP is minimal. Our method achieves the best performance when the WSGM count is set to 12 and the reduction factor is 4. Additionally, the reduction factor should not be too large, as smaller intermediate layer dimensions fail to effectively guide the frozen model to focus on forgery-specific features.

\noindent\textbf{Impact of Different FAFormer Layers.} We investigate the effect of varying the number of FAFormer layers on our approach. The results are presented in Table~\ref{tab: faformer_ablation}. The experimental results show that FAFormer achieves optimal performance when the number of layers is set to 2. Overall, the impact of different layer configurations on FAFormer’s performance is minimal.

\begin{table*}[t]
\centering
\resizebox{0.85\textwidth}{!}{
\begin{tabular}{l c c c c c c c c c}
    \toprule
    \multirow{2}{*}{Methods} & \multicolumn{8}{c}{Testing Subset} & \multirow{2}{*}{Avg. Acc.(\%)}\\
    \cmidrule(lr){2-9}
     & ADM & BigGAN & GLIDE & MidJourney & SDV1.4 & SDV1.5 & VQDM & Wukong &  \\
    \midrule
    ResNet-50~\cite{he2016deep} & 53.5 & 52.0 & 61.9 & 54.9 & 99.9 & 99.7 & 56.6 & 98.2 & 72.1 \\
    DeiT-S~\cite{touvron2021training} & 53.5 & 52.0 & 61.9 & 54.9 & 99.9 & 99.7 & 56.6 & 98.2 & 72.1 \\
    Swin-T~\cite{liu2021swin} & 49.8 & 57.6 & 67.6 & 62.1 & 99.9 & 99.8 & 62.3 & 99.1 & 74.8 \\
    CNN-Spot~\cite{wang2020cnn} & 50.1 & 46.8 & 39.8 & 52.8 & 96.3 & 95.9 & 53.4 & 78.6 & 64.2 \\
    Spec~\cite{zhang2019detecting} & 49.7 & 49.8 & 49.8 & 52.0 & 99.4 & 99.2 & 55.6 & 94.8 & 68.8 \\
    F3Net~\cite{qian2020thinking} & 49.9 & 49.9 & 50.0 & 50.1 & 99.9 & 99.9 & 49.9 & 99.9 & 68.7 \\
    GramNet~\cite{liu2020global} & 50.3 & 51.7 & 54.6 & 54.2 & 99.2 & 99.1 & 50.8 & 98.9 & 69.9 \\
    UniFD~\cite{ojha2023towards} & 71.9 & 90.5 & 85.4 & 93.9 & 96.4 & 96.2 & 81.6 & 94.3 & 88.8 \\
    NPR~\cite{tan2024rethinking} & 76.9 & 84.2 & 89.8 & 81.0 & 98.2 & 97.9 & 84.1 & 96.9 & 88.6 \\
    FreqNet~\cite{tan2024frequency} & 66.8 & 81.4 & 86.5 & 89.6 & 98.8 & 98.6 & 75.8 & 97.3 & 86.8 \\
    FatFormer~\cite{liu2024forgery} & 75.9 & 55.8 & 88.0 & 92.7 & 100.0 & 99.9 & 98.8 & 99.9 & 88.9 \\
    C2P-CLIP~\cite{tan2024c2p} & 96.4 & 98.7 & 99.0 & 88.2 & 90.9 & 97.9 & 96.5 & 98.8 & 95.8 \\
    \midrule
    ForgeLens  & 94.0 & 93.8 & 99.5 & 97.4 & 99.7 & 99.5 & 97.8 & 99.0 & \textbf{97.6} \\
    \bottomrule
\end{tabular}}
\caption{Comparison of Average Accuracy (Avg. ACC) between our method and other generated image detectors on the GenImage test sets. Each model is trained on the SDv1.4 subset and evaluated across all test sets. Accuracy is averaged over eight training cases per test set, with the top-performing results highlighted in bold.}
\label{tab: genimage}
\end{table*}

\subsection{Comparative Experiments}
To further validate the effectiveness of our method, we conducted additional comparative experiments on GenImage~\cite{zhu2023genimage}. GenImage comprises eight subsets, each containing fake images generated by a different model: AMD~\cite{dhariwal2021diffusion}, BigGAN~\cite{brock2018large}, GLIDE~\cite{nichol2021glide}, Midjourney~\cite{midjourney}, Stable Diffusion V1.4~\cite{rombach2022high}, Stable Diffusion V1.5~\cite{rombach2022high}, VQDM~\cite{gu2022vector}, and Wukong~\cite{wukong}. We used the SDv1.4 subset for training.

As shown in Table \ref{tab: genimage}, our method outperforms all existing detection approaches, achieving state-of-the-art performance with an average ACC of 97.6\%. Notably, it improves the average ACC by 8.8\% over the baseline UniFD and 1.8\% over the recent C2P-CLIP method. These results further demonstrate the effectiveness of our approach and its strong generalization in detecting forgery images from recent diffusion models. 

\begin{table}[h]
\centering
\resizebox{0.45\textwidth}{!}{
\begin{tabular}{lcc}
    \toprule
    Hyperparameter & \makecell{UniversalFakeDetect\\Value} & \makecell{GenImage\\Value} \\
    \midrule
    train\_data\_count & 1,600 & 400 \\
    train\_classes & car, cat, chair, horse & N/A \\
    \midrule
    stage1\_batch\_size  & 16 & 16 \\
    stage1\_epochs  & 50 & 10 \\
    stage1\_learning\_rate  & $5 \times 10^{-5}$ & [$1 \times 10^{-3}$, $1 \times 10^{-4}$]\\
    stage1\_lr\_decay\_step  & 2 & 3 \\
    stage1\_lr\_decay\_factor  & 0.7 & 0.7 \\
    WSGM\_count  & 12 & 4, 8, 12 \\
    WSGM\_reduction\_factor & 4 & 2, 4, 8\\
    \midrule
    stage2\_batch\_size  & 16 & 16 \\
    stage2\_epochs  & 10 & 5 \\
    stage2\_learning\_rate  & $2 \times 10^{-6}$ & $1 \times 10^{-5}$ \\
    stage2\_lr\_decay\_step  & 2 & 3 \\
    stage2\_lr\_decay\_factor  & 0.7 & 0.7 \\
    FAFormer\_layers  & 2 & 2 \\
    FAFormer\_reduction\_factor  & 1 & 1 \\
    FAFormer\_head & 2, 4 & 2, 4 \\
    \bottomrule
\end{tabular}}
\caption{Hyperparameters.}
\label{tab: hyperparameter}
\end{table}

\section{Implementation Hyperparameter Details}
\label{appendix: hyperparameter}
To facilitate the reproduction of our best results on the UniversalFakeDetect and GenImage datasets, we provide a complete list of all hyperparameters used during training, as shown in Table~\ref{tab: hyperparameter}. For the GenImage dataset, the training process exhibits some fluctuations; therefore, we provide a range of values for certain hyperparameters.

\section{Additional Class Activation Map Visualization}
\begin{figure*}[ht]
    \centering
    \includegraphics[width=0.8\linewidth]{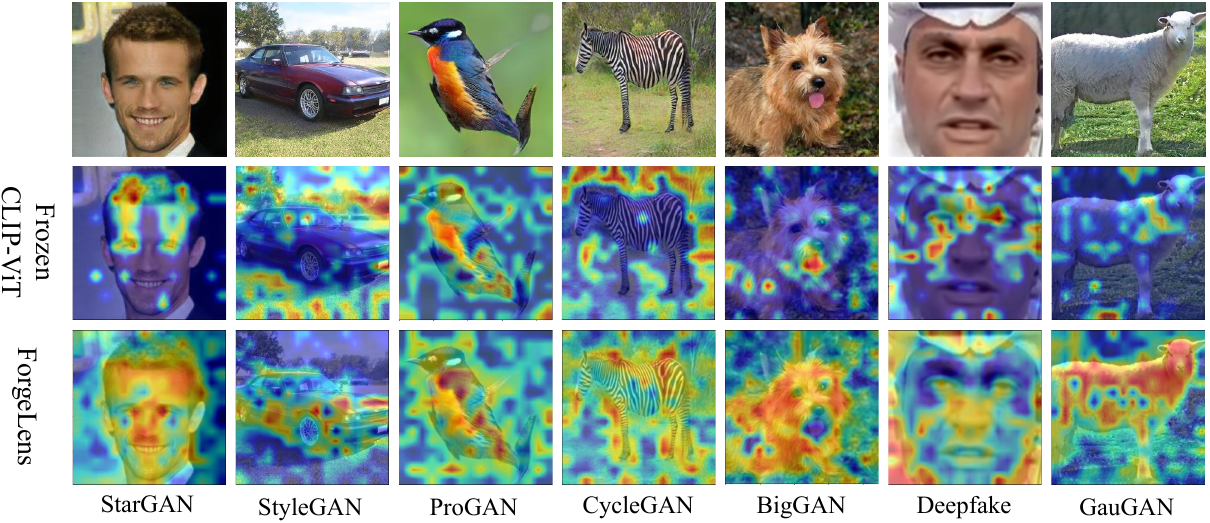}
    \caption{Class Activation Map (CAM) visualization of features extracted by the frozen CLIP-ViT
and ForgeLens on GANs data.}
    \label{fig:gam_gan}
\end{figure*}

\begin{figure*}[ht]
    \centering
    \includegraphics[width=0.9\linewidth]{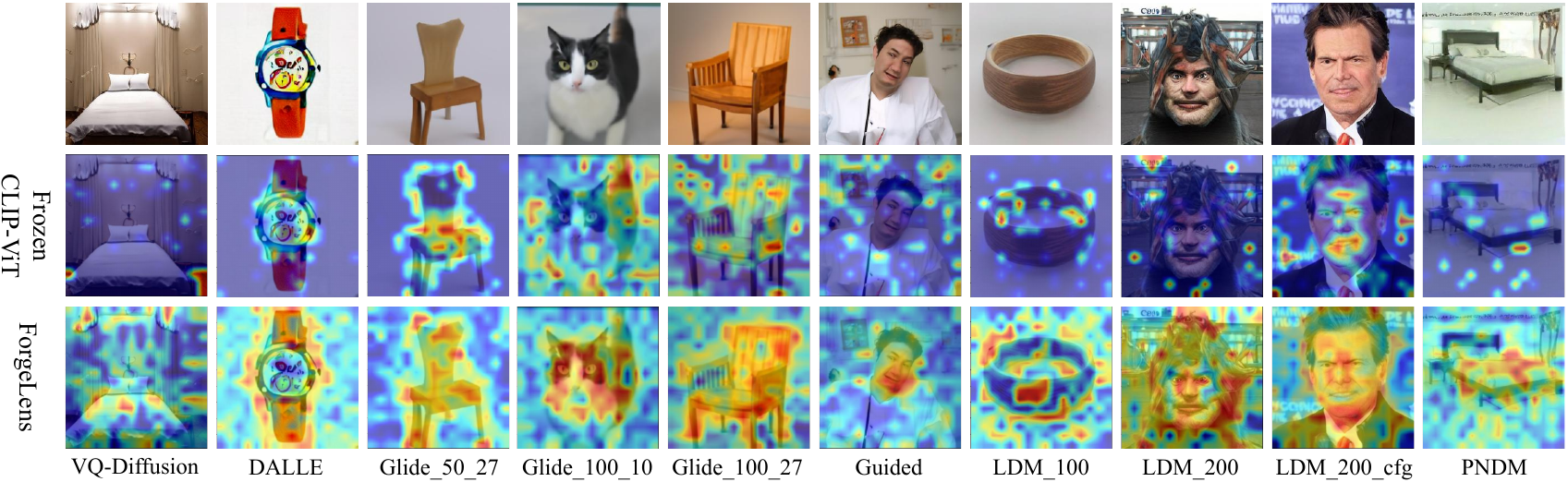}
    \caption{Class Activation Map (CAM) visualization of features extracted by the frozen CLIP-ViT
and ForgeLens on diffusion model data.}
    \label{fig:gam_diffusion}
\end{figure*}

To more comprehensively validate the effectiveness of focusing on forgery-specific features, we conducted extensive CAM visualizations on images generated by both GANs and Diffusion models, as shown in Figures~\ref{fig:gam_gan} and \ref{fig:gam_diffusion}. Compared to the general-purpose features extracted using only the frozen base model CLIP-ViT, the forgery-focused features extracted by ForgeLens exhibit stronger activations and greater attention to manipulated regions, highlighting its effectiveness in capturing forgery-related patterns.

\end{document}